\documentclass[10pt,twocolumn,letterpaper]{article}

\usepackage{cvpr}              % To produce the CAMERA-READY version
\usepackage{graphicx}
\usepackage{amsmath}
\usepackage{amssymb}
\usepackage{booktabs}

\usepackage{algorithm}
\usepackage{algpseudocode}

\usepackage[pagebackref,breaklinks,colorlinks]{hyperref}

\usepackage[capitalize]{cleveref}
\crefname{section}{Sec.}{Secs.}
\Crefname{section}{Section}{Sections}
\Crefname{table}{Table}{Tables}
\crefname{table}{Tab.}{Tabs.}
\crefname{algorithm}{Algo.}{Algos.}

\def\cvprPaperID{7946} % *** Enter the CVPR Paper ID here
\def\confName{CVPR}
\def\confYear{2022}

\begin{document}

\title{End-to-End Trajectory Distribution Prediction Based on Occupancy Grid Maps}

\author{Ke Guo$^{1}$, Wenxi Liu$^{2}$, Jia Pan$^{1}$ 
\thanks{Jia Pan is the corresponding author. This project is supported by HKSAR RGC GRF 11202119, 11207818, T42-717/20-R, HKSAR Technology Commission under the InnoHK initiative, and the National Natural Science Foundation of China (Grant No. 62072110).}\\
$^1$The University of Hong Kong \quad $^2$College of Computer and Data Science, Fuzhou University\\
{\tt\small \{kguo,jpan\}@cs.hku.hk, wenxi.liu@hotmail.com}
}
\maketitle

%%%%%%%%% ABSTRACT
\begin{abstract}
In this paper, we aim to forecast a future trajectory distribution of a moving agent in the real world, given the social scene images and historical trajectories. Yet, it is a challenging task because the ground-truth distribution is unknown and unobservable, while only one of its samples can be applied for supervising model learning, which is prone to bias. Most recent works focus on predicting diverse trajectories in order to cover all modes of the real distribution, but they may despise the precision and thus give too much credit to unrealistic predictions. To address the issue, we learn the distribution with symmetric cross-entropy using occupancy grid maps as an explicit and scene-compliant approximation to the ground-truth distribution, which can effectively penalize unlikely predictions. In specific, we present an inverse reinforcement learning based multi-modal trajectory distribution forecasting framework that learns to plan by an approximate value iteration network in an end-to-end manner. Besides, based on the predicted distribution, we generate a small set of representative trajectories through a differentiable Transformer-based network, whose attention mechanism helps to model the relations of trajectories. In experiments, our method achieves state-of-the-art performance on the Stanford Drone Dataset and Intersection Drone Dataset. 

\end{abstract}
%%%%%%%%% BODY TEXT
\section{Introduction}
\label{sec:intro}
Trajectory prediction has gained increasing attention due to its emerging applications such as robot navigation and self-driving cars. Due to the inherent multimodal uncertainty from an agent's intention or environment, a large number of works have been proposed to learn a multimodal distribution of the future trajectories. For example, in~\cite{leung2016distributional,makansi2019overcoming}, the multimodal distribution is explicitly modeled using the Gaussian mixture model, though it is hard to optimize and prone to overfitting. Others have attempted to model the trajectory distribution implicitly using generative models such as conditional variational autoencoder (CVAE)~\cite{lee2017desire,salzmann2020trajectron++,mangalam2020not,deo2020trajectory}, normalizing flow (NF)~\cite{rhinehart2018r2p2,park2020diverse}, or generative adversarial network (GAN)~\cite{gupta2018social,amirian2019social,hu2020collaborative,sadeghian2019sophie,eiffert2020probabilistic}.

\begin{figure}[t]
	\centering
	\includegraphics[width=\linewidth]{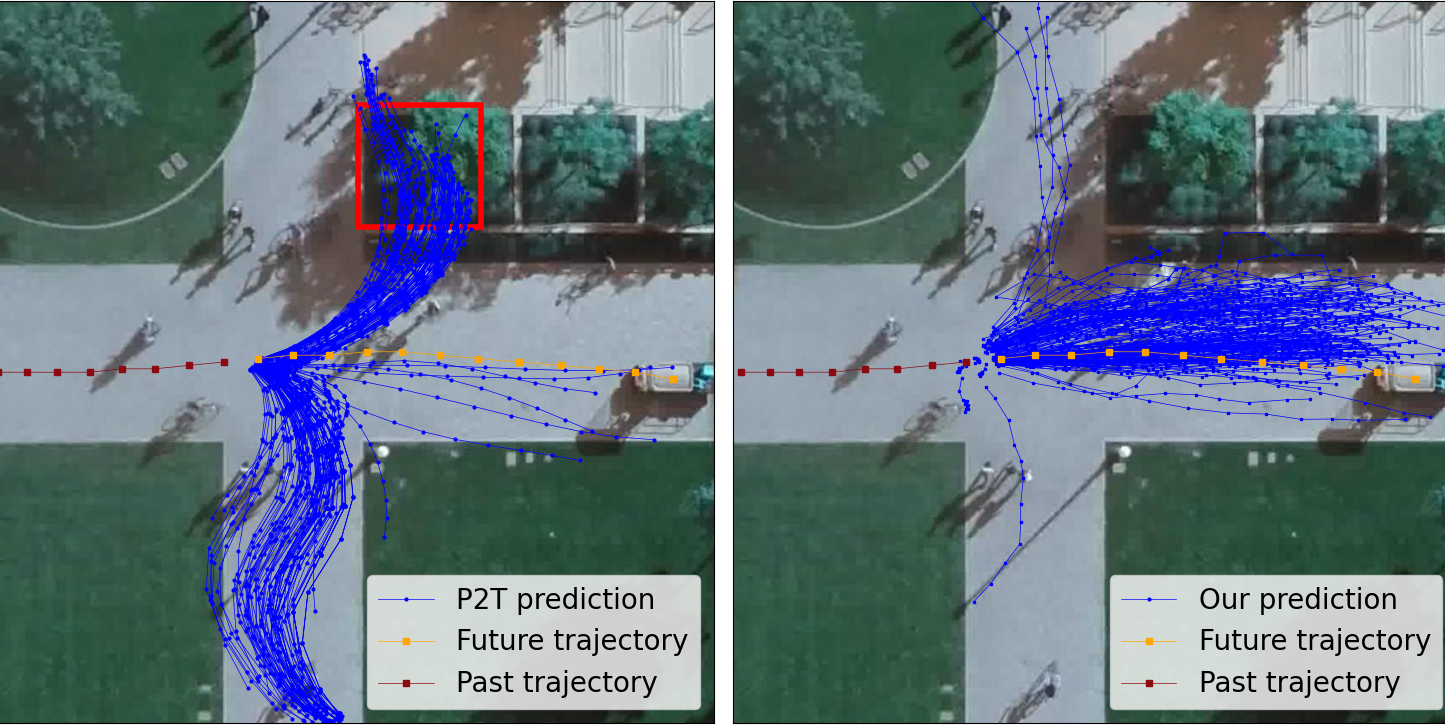}
	\vspace{-0.6cm}
	\caption{Illustration of trajectory prediction distributions of P2T and our method on Stanford Drone Dataset. Although the prediction of P2T is more diverse, it predicts many infeasible outcomes (e.g., those trajectories intersecting with the parterre) and assigns too high probability to the turning action.}
	\vspace{-0.6cm}
	\label{fig:dist}
\end{figure}

However, most previous works focus on the diversity of the predicted trajectories rather than the more important precision, except a few works (\eg~\cite{rhinehart2018r2p2,park2020diverse}). The issue is that if the model is only encouraged to cover all modes of real distribution, it may assign too many probabilities to unrealistic predictions and cannot accurately reflect the real probability density. One such example is shown in \cref{fig:dist} where a large portion of the diverse trajectories predicted by P2T~\cite{deo2020trajectory} turn and intersect with obstacles, which are certainly implausible and inconsistent with common knowledge that moving straight ahead is more likely than turning. In such circumstances, a navigation decision based on the predictions will overreact to less likely futures, while underestimating the more likely ones.

Specifically, to learn a diverse trajectory distribution, previous works usually minimize the variety loss~\cite{gupta2018social,huang2019stgat,deo2020trajectory} or the forward cross-entropy~\cite{makansi2019overcoming,salzmann2020trajectron++,liang2020garden}. Yet, the variety loss does not penalize bad predictions as long as there exists one prediction close to the ground-truth, and it does not lead to ground-truth distribution but approximately its square root~\cite{thiede2019analyzing}. On the other hand, the forward cross-entropy also fails to adequately penalize the unlikely predictions~\cite{rhinehart2018r2p2,park2020diverse} and exhibits noise sensitivity~\cite{wang2019symmetric}. To overcome the limitations of these losses, our solution is to learn a distribution minimizing the symmetric cross-entropy, \ie the combination of forward and reverse cross-entropy between the predictive distribution and ground-truth distribution. Compared with the forward cross-entropy, the reverse cross-entropy can penalize the prediction with low likelihood, but it requires ground-truth distribution as a reference, which unfortunately is not available in many cases. An effective solution is to employ an occupancy grid map (OGM), which divides the social space into grid cells with an occupancy probability in each cell. Thus, the trajectory probability can be approximated as the product of all future position probabilities conditioned on the OGM. In~\cite{rhinehart2018r2p2}, an OGM, parameterized as a cost map, is embedded from spatial scene features by a convolutional neural network (CNN) to assign proper probabilities to different social areas. However, representing all future position distributions with a single OGM is inaccurate, since it neglects the spatial-temporal correspondence of trajectories. Instead, we predict an OGM for each future position with a convolutional long short-term memory (ConvLSTM)~\cite{xingjian2015convolutional} network based on our novel deconvolution parameterization of the position probability flow. The resulting dynamic OGMs can help not only the trajectory prediction~\cite{liang2020garden} but also downstream planning tasks~\cite{zeng2019end,casas2021mp3}.

When minimizing the symmetric cross-entropy, previous approaches~\cite{rhinehart2018r2p2,park2020diverse} usually make use of the normalizing flow, which transforms a simple Gaussian distribution into the target trajectory distribution through a sequence of auto-regressive mappings. These mappings are required to be invertible, differentiable, and easy for computing Jacobian determinants, which are difficult to be satisfied in practice. In addition, the latent variable sampled from the Gaussian distribution is hard to interpret. To address these issues, we develop an end-to-end interpretable model to back-propagate the symmetric cross-entropy loss. In particular, we construct a CVAE model using a coarse future trajectory plan within neighboring grids as the interpretable latent variable, similar to P2T~\cite{deo2020trajectory}. However, P2T cannot be trained in an end-to-end manner, because it learns the planning policy using the maximum-entropy inverse reinforcement learning (MaxEnt IRL)~\cite{ziebart2008maximum,wulfmeier2017large} by matching feature expectation. Instead, we implement value iteration in IRL by differentiable value iteration network (VIN)~\cite{2016Value} and incorporate Gumbel-Softmax~\cite{jang2017categorical} into the discrete planning policy sampling. In our VIN-based IRL, planning and trajectory generation policy can be learned simultaneously by maximizing the data likelihood. 

Even though a large number of possible future trajectories can be sampled from the learned distribution, many downstream applications often demand a small set of representative predictions. This requirement is traditionally accomplished by learning the distribution model with the variety loss~\cite{gupta2018social,huang2019stgat,dendorfer2020goal} or post-processing with heuristic methods like greedy approximation~\cite{phan2020covernet} or K-means~\cite{deo2020trajectory,deo2021multimodal}. Motivated by the insight that clustering like K-means can be regarded as paying different attention to different samples, we propose a Transformer-based refinement network, whose attention mechanism can also ensure sampling diversity, to attentively obtain a small set of representative samples from the over-sampled outcomes of our prediction model.
The representative properties can be conveniently adjusted by its loss, \eg the variety loss for diversity. In experiments, we compare our method with a set of state-of-the-art approaches on the Stanford Drone Dataset~\cite{robicquet2016learning} and Intersection Drone Dataset~\cite{inDdataset} and demonstrate the superiority of our method in both trajectory diversity and quality.

In summary, the main contributions are as follows. 
\begin{itemize}
\itemsep=-5pt
    % \item We propose an end-to-end crowd trajectory forecasting framework to predict a multi-modal trajectory distribution conditioned on grid plans sampled from the policy learned by our approximate value iteration network;we propose a VIN-based IRL to bridge the gaps between discrete planning and continuous trajectory prediction, which leads to an end-to-end multi-modal trajectory prediction framework. #Our approach couples the predictions and planning. We condition the prediction process on its plan, 
    % \item We propose to predict a multi-modal trajectory distribution conditioned on grid plans sampled from the IRL policy in an end-to-end manner;
    \item We propose a VIN-based IRL method, simplifying the learning process while allowing the gradients from trajectory generation to flow back to the planning module. 
    \item We improve the approximation of ground-truth with OGMs in
    learning trajectory distribution using symmetric cross-entropy.;
    % Our model is enabled to learn a diverse and feasible trajectory distribution by minimizing the symmetric cross-entropy with the OGMs as an approximation to the ground-truth distribution, and the probability density transition process of successive OGMs can be well modeled by deconvolution.
    % \item predicting the OGMs using the deconvolution operation to directly model the probability density transition process between the successive OGMs;
    \item We introduce a Transformer-based refinement network for sampling from trajectory distribution to obtain representative and realistic trajectories; 
    \item We demonstrate the state-of-the-art performance of our framework on two real-world datasets: Stanford Drone dataset~\cite{robicquet2016learning} and Intersection Drone dataset~\cite{inDdataset}.
\end{itemize}

\section{Related Work}
\label{sec:related}

\subsection{Trajectory Distribution Prediction}

We focus on trajectory distribution prediction approaches based on deep learning. Refer to~\cite{rudenko2020human} for a survey of more classical approaches. In the early literature, the trajectory distribution is usually modeled as a simple unimodal distribution, \eg a bivariate Gaussian distribution~\cite{alahi2016social,luo2018fast,zhang2019sr}. However, the unimodal models tend to predict the average of all possible modes, which may be not valid.

Recently, various generative models such as GAN, NF and CVAE, have been proposed to address the multi-modality, which capture the stochasticity with a latent variable. GAN-based methods~\cite{gupta2018social,amirian2019social,hu2020collaborative,sadeghian2019sophie,eiffert2020probabilistic} use a discriminator to generate diverse realistic trajectories but are difficult to train and suffer from the mode collapse. NF methods~\cite{rhinehart2018r2p2,park2020diverse} sample the latent variable from a standard Gaussian distribution and map it to the target trajectory through a sequence of transformations. Some CVAE approaches such as DESIRE~\cite{lee2017desire}, Trajectron++\cite{salzmann2020trajectron++}, learn a Gaussian or categorical latent distribution using the constraint between prior and posterior distribution. Others leverage an interpretable latent variable to incorporate prior knowledge. For example, PECNet~\cite{mangalam2020not}, TNT~\cite{zhao2020tnt} consider the destination position as the latent variable. Further, LB-EBM~\cite{pang2021trajectory} takes positions at several future steps as the latent variable which is sampled from an energy-based model. P2T~\cite{deo2020trajectory} samples a coarse grid plan from a policy learned by deep MaxEnt IRL~\cite{wulfmeier2017large} as the latent variable. Even though our model also leverages the plan as the latent variable, we learn the plan and trajectory distributions in a unified framework.

\subsection{Occupancy Grid Maps Prediction}

OGMs prediction works aim at predicting the categorical occupancy distribution over a grid at each future time step. Even though there are extensive studies forecasting OGMs of a crowd~\cite{li2017pedestrian,minoura2020crowd} or all objects~\cite{Mohajerin2019MultiStep,luo2021safety} in a scene, we focus on reviewing the literature predicting one agent's OGMs like our work.  

Kim \etal~\cite{kim2017probabilistic} directly outputs the future probability of each grid cell using the LSTM network. Y-net~\cite{mangalam2021goals} yields the OGM at each future step directly from different channels of the feature map output by a CNN. Similarly, in MP3~\cite{casas2021mp3}, the feature map at each channel is embedded into the temporal motion fields at each future step, which obtains the probability transition flow between consecutive OGMs by bilinear interpolation of motion vectors on the field. To take advantage of the temporal and spatial patterns in the sequential OGMs, ConvLSTM~\cite{xingjian2015convolutional} is widely applied. In~\cite{ridel2020scene}, they directly derive an OGM from the hidden map of ConvLSTM at each time step. To increase time consistency, DRT-NET~\cite{jain2020discrete} learns the residual provability flow between the consecutive OGMs. To incorporate the prior knowledge of local movement, Multiverse~\cite{liang2020garden} uses a graph attention network to aggregate neighborhood information on the hidden map of ConvLSTM. Similarly, SA-GNN~\cite{luo2021safety} considers the interactions with neighbors by graph neural networks. Based on the ConvLSTM and deconvolution parameterization, our method is not only computationally efficient but also explicitly models the local transition probability.

Furthermore, some of these works attempt to obtain trajectories by sampling the OGMs. But the positions independently sampled from each OGM suffer from discretization errors and lack spatio-temporal correspondence in trajectories. To address this problem, \cite{ridel2020scene} leverages the OGMs as input to another ConvLSTM which outputs the coordinates of a fixed number of future trajectories. Multiverse~\cite{liang2020garden} predicts a continuous offset at each cell to mitigate the discretization error and applies a diverse beam search to generate multiple distinct trajectories. Y-net~\cite{mangalam2021goals} samples intermediate positions conditioned on the sampled goal and waypoints. Instead of sampling OGMs like all previous works, we use the OGMs as auxiliary information in training loss to generate more feasible trajectories.

\subsection{Trajectory Sample Refinement}

Trajectories sampled from the predicted trajectory distribution usually do not satisfy the downstream requirement. The most common two requirements are precision and diversity to cover all future scenarios accurately~\cite{rhinehart2018r2p2,park2020diverse}. To improve accuracy, previous works~\cite{lee2017desire,zhao2020tnt,marchetti2020mantra} usually score the samples using a neural network and refine the top samples. For diversity, the relation between samples needs to be considered. Most literature~\cite{gupta2018social,huang2019stgat,dendorfer2020goal} directly use a variety loss to improve diversity. In addition, P2T~\cite{deo2020trajectory}, PGP~\cite{deo2021multimodal} and Y-net~\cite{mangalam2021goals} use K-means to cluster samples while CoverNet~\cite{phan2020covernet} employs a greedy approximation algorithm to create a diverse set. To capture both diversity and quality, DSF~\cite{yuan2020diverse} learns a diverse sampling function to sample the latent variable of CVAE using a diversity loss based on a determinantal point process at test time while DiversityGAN~\cite{huang2020diversitygan} samples distinct latent semantic variables to predict diverse trajectories. Different from previous work, our sample refinement network based on Transformer is an independent and differentiable module and is flexible with the downstream requirement and trajectory sample number.

\section{Background}
\label{sec:background}
\subsection{Problem Formulation}

Given an observation $\Omega$ including a context and history trajectory $X=\{X_t\in \mathbb{R}^2 \mid t=-t_p+1,\dots,0\}$ of a target agent, our objective is to predict the distribution $p(Y|\Omega)$ of its future trajectory $Y=\left\{Y_t\in \mathbb{R}^2\mid t=1,\dots,t_f\right\}$. The context consists of neighbors' history trajectories and an image $\mathbf{I}$, which is a bird's eye view (BEV) perception of the local scene centered at the agent's current position.

We assume that an agent has a grid-based plan on which its future trajectory is conditioned. An agent's planning process is modeled using a Markov decision process (MDP) $\mathcal{M}=\{\mathcal{S}, \mathcal{A}, \mathcal{T}, \mathbf{r}\}$, with a time horizon $N$. A state set $\mathcal{S}$ consists of all cells over a 2D grid and an absorbing end state of zero value. An action set $\mathcal{A}$ includes 4 adjacent movements ${up, down, left, right}$ and an $end$ action leading to the absorbing state. A deterministic transition function $\mathcal{T}: \mathcal{S} \times \mathcal{A} \rightarrow \mathcal{S}$ describes system dynamics. A non-stationary reward function $\mathbf{r}^n:\mathcal{S} \times \mathcal{A} \rightarrow \mathbb{R}$ determines a reward for each state and action per step $n$. We assume that the agent uses a non-stationary stochastic policy $\boldsymbol{\pi}^n(a|s)$ to determine the probability of selecting an action $a$ at a state $s$ at MDP step $n$, and finally make a plan in terms of the state sequence $S=\{s^n \in \mathcal{S}\mid n=1,\dots,N\}$. Note that here we are using the superscript $n$ as the MDP step $n$, to distinguish with the time step $t$ as subscript.

To relieve the difficulty of modeling the multi-modal future trajectory distribution $p\left(Y|\Omega\right)$, we introduce the plausible plan as the latent variable and decompose it as:
\begin{equation}
p\left(Y|\Omega\right)=\int_{S \in \mathbb{S}(\Omega)} p\left(S|\Omega\right) p\left(Y|S, \Omega\right) d S,
\label{eq:decompose}
\notag
\end{equation}
where $\mathbb{S}(\Omega)$ is the space of plausible plans conditioned on the observation. In this way, since the plan uncertainty can well capture the multimodality, trajectory conditioned on a plan can be well approximated as a unimodal distribution.

\begin{figure*}[t]
	\centering
	\includegraphics[width=\linewidth]{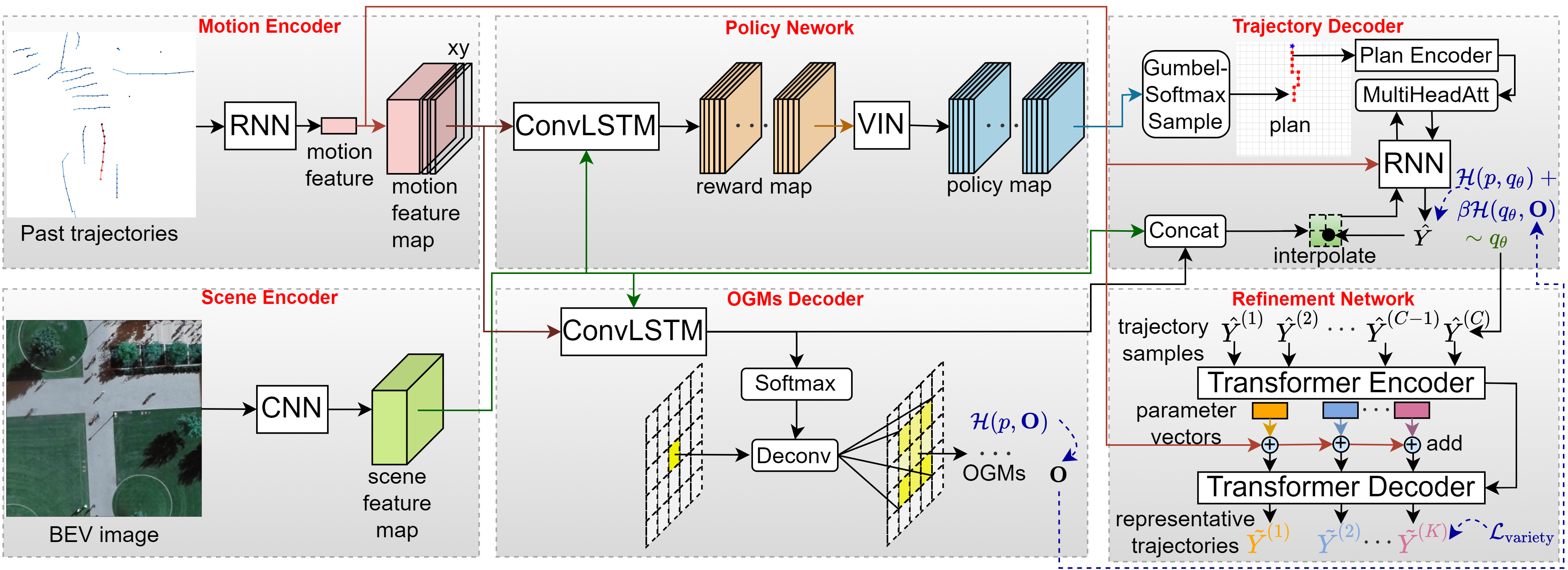}
	\vspace{-0.7cm}
	\caption{Overview of our approach.}
	\vspace{-0.5cm}
	\label{fig:overview}
\end{figure*}

\subsection{Trajectory Distribution Learning}

We predict the future trajectory distribution by minimizing the discrepancy between the distribution $q_\theta(\hat{Y}|\Omega)$ of the predicted trajectory $\hat{Y}$ and the ground-truth distribution $p(Y|\Omega)$. As a straightforward distance metric between these two distributions, forward cross-entropy (a.k.a, negative log-likelihood (NLL)) is computed as:
\begin{equation}
\begin{aligned}
\mathcal{H}\left(p, q_{\theta}\right)=&-\mathop{\mathbb{E}}\limits_{\Omega \sim \Psi,Y \sim p(\cdot|\Omega), S \in \mathbb{S}(Y)} \left[\log  q_{\theta}(S|\Omega)q_{\theta}(Y|\Omega, S)\right], \notag
\end{aligned}
\end{equation}
where $\Psi$ denotes the ground-truth observations' distribution and $\mathbb{S}(Y)$ is the space containing the ground-truth plan $S$, \ie the grid state sequence the trajectory $Y$ goes through.

Although the NLL loss encourages the predicted distribution to cover all plausible modes of the ground-truth distribution, it assigns a low penalty to the implausible predictions which are less likely to take place under the ground-truth distribution~\cite{rhinehart2018r2p2,park2020diverse}. The reverse cross-entropy $\mathcal{H}(q_\theta,p)$ can evaluate the likelihood of the prediction under the ground-truth distribution and penalize unlikely predictions, but the ground-truth distribution $p$ is unknown in the real world with only one sample observed. To address this issue, we approximate the continuous joint distribution $p(Y|\Omega)$ of future trajectory as a product of future positions' categorical marginal distributions $\mathbf{O}=\{\mathbf{O}_t\mid t=1,\dots,t_f\}$, represented as OGMs:
\begin{equation}
    p(Y|\Omega) \approx p(\mathbf{O}|\Omega) \prod \limits_{t=1}^{t_f} \mathbf{O}_t(Y_t) , \notag
\end{equation}
where $\mathbf{O}_t(Y_t)$ denotes the agent's location probability at $Y_t$ at time $t$, which is bilinearly interpolated from nearby probabilities on $\mathbf{O}_t$ and $p(\mathbf{O}|\Omega)$ is assumed to be deterministic and parameterized by neural networks $\mathbf{O}=o_\alpha(\Omega)$. Thus, the reverse cross-entropy $\mathcal{H}(q_\theta,p)$ can be approximated as:
\begin{equation}
\mathcal{H}(q_\theta,\mathbf{O}) = -\mathop{\mathbb{E}}\limits_{\Omega \sim \Psi,\hat{Y} \sim q_\theta(\cdot|\Omega)}\log p(\mathbf{O}|\Omega)\prod  \limits_{t=1}^{t_f} \mathbf{O}_t(\hat{Y}_t). \notag
\end{equation}

\section{Approach}
\label{sec:approach}
As shown in \cref{fig:overview}, our model is composed of five modules that can be learned in an end-to-end manner: an \textbf{Observation Encoder}, a \textbf{Policy Network}, an \textbf{Occupancy Grid Maps Decoder (OGMs Decoder)}, a \textbf{Trajectory Decoder} and a \textbf{Refinement Network}.

% As shown in \cref{fig:overview}, our model is composed of five modules that can be learned in an end-to-end manner: an \textbf{Observation Encoder} consisting of Scene Encoder and Motion Encoder to encode the observation into features; a \textbf{Policy Network} to generate a non-stationary policy based on the observation features; an \textbf{Occupancy Grid Maps Decoder (OGMs Decoder)} to predict the future OGMs using the observation features; a \textbf{Trajectory Decoder} to produce a future trajectory distribution conditioned on a plan; and a \textbf{Refinement Network} to yield a small set of predictions based on a large number of trajectory samples.

\subsection{Observation Encoder}

The first component of our approach is an observation encoder composed of a motion encoder to extract motion features from the past trajectories of the target and its neighbors and a scene encoder to extract scene features from the BEV image of the surrounding environment. 

\noindent{\textbf{Motion encoder:}} The motion encoder is designed to embed the past trajectories of the target agent and its neighbors into a feature vector and a feature map. To represent the neighbors' state succinctly, we leverage a directional pooling grid from~\cite{kothari2021human}, where each cell contains the relative velocity of a neighbor located in that cell with respect to the target agent. At each past time step $t$, we first flatten the grid into a vector $d_t$ and then concatenate the vector with the agent velocity $X_t-X_{t-1}$ as input to an RNN. The hidden state of the RNN at time $t$ is given by:
\begin{equation}
m_{t} =\operatorname{RNN_m}\left(m_{t-1}, \phi \left[d_{t},X_t-X_{t-1}\right]\right), \notag
\end{equation}
where $\phi$ is a linear embedding layer and the brackets indicate concatenation. The first hidden state $m_{-t_p+1}$ is set to zero and the last hidden state ${m_0}$ is regarded as the motion feature. The $m_0$ is duplicated over all cells in the scene and then is concatenated with each cell's agent-centered, world-aligned coordinate to construct a motion feature map  $\mathbf{M}$:
\begin{equation}
\mathbf{M}(x,y) =[m_0,x,y]. \notag
\end{equation}

\noindent{\textbf{Scene encoder:}} We apply a CNN to extract a scene feature map from the BEV image $\mathbf{I}$ of the neighborhood:
\begin{equation}
\mathbf{F}=\operatorname{CNN_f}(\mathbf{I}), \notag
\end{equation}
where the spatial dimensions of the scene feature map $\mathbf{F}$ are the same as that of the MDP grid for simplicity. 

\subsection{Policy Network}

We generate a policy in two steps end-to-end: mapping the observation features into rewards and then computing a policy with a value iteration network. 

We adopt non-stationary rewards to capture the dynamic agent-to-scene and agent-to-agent interaction. Based on the scene and motion feature maps, a ConvLSTM architecture is applied to yield the reward map at each step. The ConvLSTM hidden map and the reward map at MDP step $n$ are:
\begin{equation}
\mathbf{H}^n=\operatorname{ConvLSTM_r}(\mathbf{H}^{n-1},\mathbf{F}),~~~\mathbf{r}^n=\Phi(\mathbf{H}^n), \notag
\end{equation}
where $\Phi$ is a fully connected convolutional layer. The initial hidden map $\mathbf{H}^0$ is the embedded motion feature map $\Phi(\mathbf{M})$. 

Based on the reward maps, we use the approximate value iteration to generate a policy map $\boldsymbol{\pi}^n$ at each step $n$. To back-propagate the loss through the value iteration, we take advantage of the value iteration network as~\cite{2016Value,rehder2018pedestrian, pflueger2019rover}, which recursively computes the next value map by a convolution of the current value map with transition filters. To improve the value iteration network's performance, we utilize an approximate value iteration in the MaxEnt IRL formulation~\cite{ziebart2008maximum,wulfmeier2017large} with non-stationary rewards. \cref{algo:Value} describes the overall computation process of this network.

\setlength{\textfloatsep}{0.1cm}
\begin{algorithm}[t]
	\begin{algorithmic}[1]
		\Require  $\mathbf{r}^n(s,a)$
		
		\Ensure  $\boldsymbol{\pi}^n(a|s)$
	
		\State $\mathbf{V}^N(s)=0, \forall s \in \mathcal{S}$;
		
		\For{$n=N,\dots,2,1$}

		\State $\mathbf{Q}^n(s,a)=\mathbf{r}^n(s,a)+\mathbf{V}^n_{s^{\prime}=\mathcal{T}(s, a)}(s^\prime)$, 		 $\forall s \in \mathcal{S}$, $\forall a \in \mathcal{A}$;
		\State $\mathbf{V}^{n-1}(s)=\operatorname{logsumexp}_a \mathbf{Q}^n(s,a)$, $\forall s \in \mathcal{S}$;
		\State $\boldsymbol{\pi}^n(a|s)=\operatorname{softmax}_a\mathbf{Q}^n(s,a)$, $\forall s \in \mathcal{S}$;
		\EndFor
	\end{algorithmic}
	\caption{Approximate Value Iteration Network}
	\label{algo:Value} 
\end{algorithm}
\setlength{\floatsep}{0.1cm}

% \begin{figure}[t]
% 	\centering
% 	\includegraphics[width=0.8\linewidth]{vi.png}
% 	\caption{Approximate Value Iteration Network. The network applies the value iteration in CNN representation to compute a non-stationary policy based on a non-stationary reward.}
% 	\label{fig:vi}
% \end{figure}

%,     $\mathbf{V}^\prime=\operatorname{Conv}(\mathbf{V})$
\subsection{OGMs Decoder}

To provide an explicit approximation of the ground-truth trajectory distribution, we predict a sequence of dynamic OGMs based on the observation features using a ConvLSTM network. With the scene feature map as input, the hidden map of the ConvLSTM network at time $t$ is:
\begin{equation}
\mathbf{H}_t=\operatorname{ConvLSTM_o} \left(\mathbf{H}_{t-1},\mathbf{F}\right), \notag
\end{equation}
The hidden map is initialized with the embedding of the motion feature map $\mathbf{H}_0=\Phi(\mathbf{M})$.

Then, instead of directly outputting an OGM from each hidden map, we derive a pixel-adaptive normalized deconvolution filter whose weights are spatially varying, non-negative and sum to one. The deconvolution is subsequently applied to the last OGM to obtain the next one:
\begin{equation}
\mathbf{O}_{t}=\operatorname{Deconv} \left(\mathbf{O}_{t-1},\operatorname{softmax}(\Phi(\mathbf{H}_t))  \right), \notag
\end{equation}
where the initial OGM $\mathbf{O}_{0}$ is a probability matrix to be learned. Our deconvolution method can directly model the probability density transition process. Besides, the limited size of the normalized deconvolution kernel ensures that the probability mass diffuses into nearby grid cells in a conservative manner, reflecting the prior knowledge that agents do not suddenly disappear or jump between distant locations.

% \subsection{Plan Sampling}

%\subsection{Trajectory Generation}
%and the trajectory decoder output
%  $\hat{Y}^{(i)}$
\subsection{Trajectory Decoder}

Conditioned on a plan from the policy roll-out or the data, an RNN decoder is applied to generate the future position distribution recursively based on local features.
% \begin{equation}
%     P(Y|S,\Omega)=\prod  \limits_{t=1}^{t_f}   P(Y_t|Y_{t-1},\dots,Y_{1},S,\Omega).
% \end{equation} from the policy roll-out or a expert plan

\noindent{\textbf{Plan sampling:}} We generate a plan $\hat{S}=\{\hat{s}^n \in \mathbb{R}^2 \mid n=1,\dots,N\}$ by sampling the non-stationary policy outputted by the policy network. However, directly sampling the policy with discrete state and action spaces will introduce difficulty in loss back-propagation. To overcome this difficulty, we sample the policy with the Gumbel-Softmax trick~\cite{jang2017categorical}, resulting in continuous action and state. Besides, we obtain the policy at continuous state $\hat{s}^n$ by bilinear interpolation.

\begin{figure}[!h]
	\centering
	\includegraphics[width=0.7\linewidth]{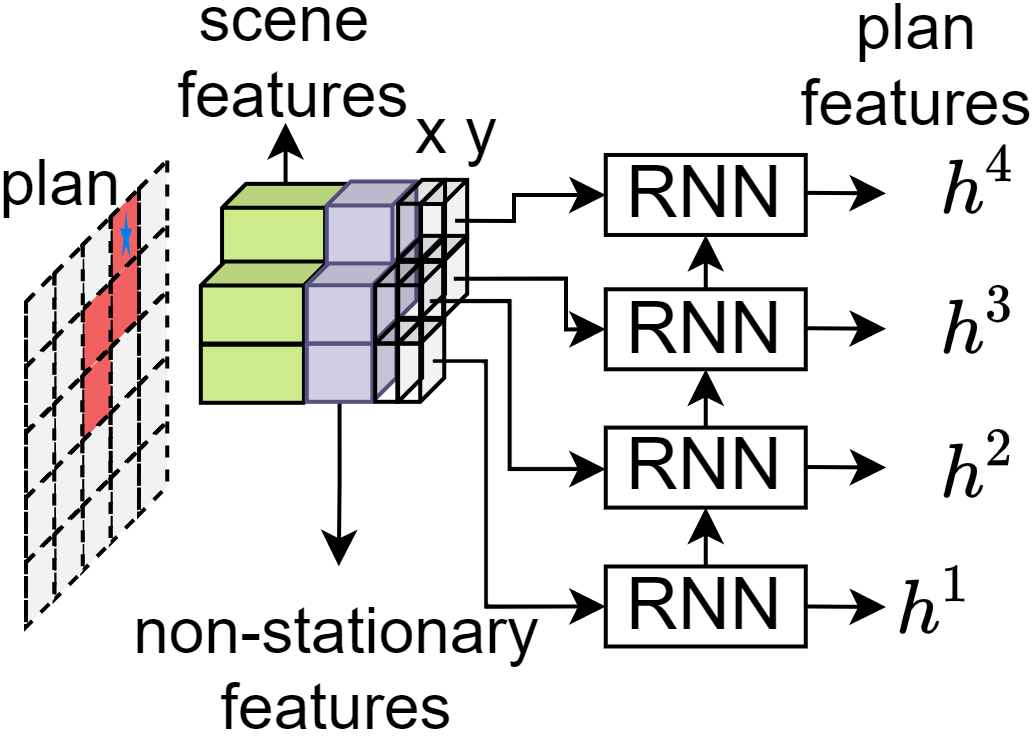}
	\vspace{-0.2cm}

	\caption{The local scene and non-stationary features at each plan state are concatenated with its location coordinates and then fed into an RNN to obtain all plan features.}

	\label{fig:plan}
\end{figure}

\noindent{\textbf{Plan encoder:}} Given a ground-truth plan $S$ (or a sampled plan $\hat{S}$), we first collect the local scene feature from scene feature map $\mathbf{F}$ and non-stationary feature from the corresponding hidden map of $\operatorname{ConvLSTM_r}$ at each plan state. Then we concatenate these features with the state's coordinates as input to an RNN, whose hidden state at step $n$ is:
\begin{equation}
h^{n} =\operatorname{RNN_s}\left(h^{n-1},\phi\left[{s}^n,\mathbf{F}({s}^n),\mathbf{H}^n({s}^n)\right]\right). \notag
\end{equation}
Since the sampled plan's state $\hat{s}^n$ is on the continuous plane, the local features like $\mathbf{F}({s}^n)$ are gathered by bilinear interpolation at the spatial dimensions of the feature map $\mathbf{F}$ corresponding to the physical position ${s}^n$. \cref{fig:plan} illustrates how the plan encoder extracts the plan features $h^{1:N}=\{h^n\mid n=1,\dots,N\}$.

\noindent{\textbf{Multi-head attention based decoder:}} Since different dimensions of the plan features at different steps may have different impacts on current hidden state~\cite{mercat2020multi}, we utilize a multi-head scaled dot product attention module~\cite{vaswani2017attention} to aggregate the plan information:
\begin{equation}
\begin{aligned}
&\operatorname{MultiHead}(Q,K,V)=[{\operatorname{Att}(QW_i^Q,KW_i^K,VW_i^V)}_{i=1}^H]W^O,\\
&\text{where} \operatorname{Att}(Q_i,K_i,V_i)=\operatorname{softmax}(\frac{Q_iK_i^T}{\sqrt{d_k}})V_i,\notag
\end{aligned}
\end{equation}
where $d_k$ is the dimension of each head. At each future time $t$, we linearly project the trajectory decoder's previous hidden state $h_{t-1}$ into the query $Q_i$ and the plan features into the key $K_i$ and value $V_i$ through linear layers $W_i^Q$, $W_i^K$ and $W_i^V$. The attention module output $a_t$ is then concatenated with coordinates and local bilinearly interpolated features on scene feature map and corresponding OGM hidden map at the previous position $Y_{t-1}$ as input to an RNN decoder:
\begin{equation}
\begin{aligned}
a_t&=\operatorname{MultiHead}(h_{t-1},h^{1:N},h^{1:N}),\\
h_t&=\operatorname{RNN_t}\left(h_{t-1}, \phi\left[a_t,Y_{t-1},\mathbf{F}(Y_{t-1}),\mathbf{H}_t(Y_{t-1})\right]\right), \notag
\end{aligned}
\end{equation}
where the initial hidden state $h_0$ is the embedded motion feature $\phi(m_0)$. Then, the hidden state $h_t$ is utilized to predict the position $\hat{Y}_t$ distribution which is assumed to be a bivariate Gaussian distribution parameterized with the mean $\mu_t+Y_{t-1}$, standard deviation $\sigma_t$, and correlation $\rho_t$:
\begin{equation}
    [\mu_t,\sigma_t,\rho_t]=h_tW^P,~~\hat{Y}_t \sim \mathcal{N}(\mu_t+Y_{t-1},\sigma_t,\rho_t). \notag
\end{equation}
During generating predictions $\hat{Y}$, the above ground-truth position $Y_{t-1}$ is substituted by the position $\hat{Y}_{t-1}$ sampled from the predicted distribution with the reparameterization trick~\cite{Kingma2014Auto} to ensure differentiability.

\subsection{Refinement Network}

We design a refinement network to present a succinct representation of a trajectory distribution with several representative trajectories. The network is an encoder-decoder framework based on Transformer~\cite{vaswani2017attention} but without positional embedding and auto-regressive decoding because the multi-head attention module in Transformer can well capture the relation between unordered samples to ensure diversity. We first over-sample a large number of trajectory samples $\{\hat{Y}^{(1)},\hat{Y}^{(2)},\dots,\hat{Y}^{(C)}\}$ to cover the trajectory distribution, \eg $C=200$. Then, all trajectory samples are flattened into vectors and embedded as input to the Transformer encoder without the positional embedding. To save inference time, we utilize a generative style decoder like~\cite{zhou2021informer} but the inputs to our decoder are the summations of the embedded motion features and $K$ different parameter vectors instead of fixed tokens. Finally, we embed the decoder output to obtain a few representative trajectories $\{\tilde{Y}^{(1)},\tilde{Y}^{(2)},\dots,\tilde{Y}^{(K)}\}$, \eg $K=20$. 

\subsection{Training Process}

To achieve different goals including a good OGM, distribution and representative sets at different steps, our training process has the following four steps:

\begin{enumerate}
\vspace{-0.2cm}
\itemsep=-5pt
\item \textbf{OGMs learning:} The observation encoder and OGMs decoder are trained to predict OGMs by minimizing the NLL loss:
\begin{equation}
\mathcal{H}(p,\mathbf{O})=-\mathop{\mathbb{E}}\limits_{\Omega \sim \Psi,Y \sim p(\cdot|\Omega),\mathbf{O}=o_\alpha(\Omega)}\log \prod  \limits_{t=1}^{t_f}\mathbf{O}_t(Y_t). 
\label{eq:ogm-learning-loss}
\notag
\end{equation}

\item \textbf{Trajectory distribution learning:} Based on the learned observation encoder and OGMs decoder, we train the policy network and trajectory decoder to induce a trajectory distribution that minimizes the approximated symmetric cross-entropy loss:
\begin{equation}
\mathcal{L}_{\text{sce}}=\mathcal{H}(p,q_\theta)+\beta \mathcal{H}(q_\theta,\mathbf{O}). \notag
\label{eq:sce}
\end{equation}

\item \textbf{Representative trajectories learning:} Using the trajectories sampled from the learned distribution, we train the refinement network to generate representative trajectories with the variety (MoN) loss~\cite{gupta2018social}:
\begin{equation}
    \mathcal{L}_{\text{variety}}=\min_{k\in\{1,\dots,K\}}\|Y-\tilde{Y}^{(k)}\|_2. \notag
\end{equation}

\item \textbf{End-to-end fine-tuning:} We fine-tune the whole network in an end-to-end manner with the variety loss.
% \begin{equation}
%      \mathcal{L}=\mathcal{H}(p,\mathbf{O})+\mathcal{L}_{\text{sce}}+\mathcal{L}_{\text{variety}}. \notag
% \end{equation}
\vspace{-0.2cm}

\end{enumerate}

Only the first two steps are required for learning a trajectory distribution while all four steps are for obtaining a compact set of representative trajectories.

\section{Experimental Results}
\label{sec:experiment}
%  $40\mathrm{m}\times40\mathrm{m}$
\subsection{Implementation Details}
We augment all trajectories and scene images in the training data by $90^{\circ}$ rotations and flipping. All data fed into and generated from our model are in the world coordinates rather than the pixel coordinates as in previous works~\cite{Sadeghian2018Trajnet,deo2020trajectory,mangalam2020not}. The BEV image fed to the scene encoder is a $200\times200$ crop of the RGB image around the agent's location. The scene encoder $\operatorname{CNN}_f$ consists of the first two layers of ResNet34~\cite{he2016deep} and a convolutional layer with kernel size 2 and stride 2, which outputs scene feature map of 32 channels and size $25\times25$ as the MDP grid. The RNNs are implemented by gated recurrent units (GRU) with hidden size 64. $\operatorname{ConvLSTM_r}$ and $\operatorname{ConvLSTM_d}$ have 1 and 2 layers each, with kernel size 3 and 32 hidden states respectively, and the deconvolution kernel size is 5. The multi-head attention module in the trajectory decoder has 4 heads of 16 dimensions. The Transformer encoder and decoder consist of 3 layers with hidden size 64 and 8 self-attention heads and a dropout rate of 0.1. We train using Adam optimizer with a learning rate of 0.001 in the first three steps and 0.0001 in the last step. We have released our code at \url{https://github.com/Kguo-cs/TDOR}. 

\subsection{Datasets and Metrics}

\textbf{Datasets.} We evaluate our method on two datasets. Most of our tests are conducted on the
\textbf{Stanford Drone Dataset} (SDD)~\cite{robicquet2016learning} provides top-down RGB videos captured on the Stanford University campus by drones at 60 different scenes, containing annotated trajectories of more than 20,000 targets such as pedestrians, bicyclists and cars. Early works~\cite{sadeghian2019sophie,dendorfer2020goal,liang2020garden} consider all trajectories in SDD and subsequent works~\cite{marchetti2020mantra,mangalam2020not,mangalam2021goals,zhao2021you} focus on pedestrian trajectories using the \textit{TrajNet} benchmark~\cite{Sadeghian2018Trajnet}. On these two splits,
%%(using the data from~\cite{deo2020trajectory} and~\cite{mangalam2020not} respectively)
we report the results of predicting the 12-step future with the 8-step history with 0.4 seconds step interval. Besides, we report our long-term prediction results on the \textbf{Intersection Drone Dataset} (inD)~\cite{inDdataset}, which consists of longer drone recorded trajectories of road users than SDD collected at four German intersections. To evaluate our method's long-term forecasting performance, we use data in~\cite{mangalam2021goals}, including 1222 training and 174 test trajectories with 5-second history and 30-second future with the 1Hz sampling rate.  
% \textbf{Intersection Drone Dataset} (inD)~\cite{inDdataset}, which consists of drone recorded trajectories of road users collected at four German intersections, and provides longer trajectories than SDD. To evaluate our method's long-term forecasting performance, we use data provided in~\cite{mangalam2021goals}, including 1222 training and 174 test trajectories with 5-second history and 30-second future with the 1Hz sampling rate.  

\textbf{Metrics. } We evaluate our performance of representative samples with three metrics. The first two are commonly used sample-based diversity metrics~\cite{gupta2018social}: $\operatorname{minADE}_{K}$, \ie, the \textit{minimum average}, and $\operatorname{minFDE}_{K}$, \ie, the \textit{final displacement errors} between $K$ predictions and ground-truth trajectory in pixels. Following P2T~\cite{deo2020trajectory}, we also report results on the quality metric, \textit{Offroad Rate}, which measures the fraction of the predicted positions falling outside road while ground-truth positions inside road. 

\begin{table}
    \centering
    % \footnotesize
    \resizebox{0.95\linewidth}{!}{
	\begin{tabular}{lccc}
    \toprule
	Model & $\operatorname{minADE}_{20}$ & $\operatorname{minFDE}_{20}$ & Offroad Rate\\
    \midrule
    S-GAN~\cite{gupta2018social}	 & 27.25  & 41.44 & -\\
	Desire~\cite{lee2017desire} & 19.25 & 34.05 & -\\
	Multiverse~\cite{liang2020garden} & 14.78 & 27.09 & -\\
	SimAug~\cite{liang2020simaug} & 10.27 & 19.71 & - \\
	P2T~\cite{deo2020trajectory} & 10.97 & 18.40 & 0.065 \\
	Ours & \textbf{8.60} & \textbf{13.90} & \textbf{0.050} \\
	\midrule
	{PECNet}~\cite{mangalam2020not} & 9.96 & 15.88 & 0.071 \\
	{LB-EBM}~\cite{pang2021trajectory} & 8.87 & 15.61 & 0.070 \\
	{P2T}~\cite{deo2020trajectory} & 8.76 & 14.08 & - \\
    {Y-Net}~\cite{mangalam2021goals} & 7.85 & 11.85 & \textbf{0.048} \\
	{V}~\cite{wong2021view} & 7.34 & 11.53 & - \\
	{Ours} & \textbf{6.77} & \textbf{10.46} & 0.066 \\
    \bottomrule
	\end{tabular}}
	\vspace{-0.2cm}
	\caption{Comparison with state-of-the-art methods on the entire SDD dataset (above) and its \textit{TrajNet} split (below) in predicting short-term 4.8-second future.}

	\label{tab:result}
\end{table}

\subsection{Performance Evaluation}

We benchmark against the following state-of-the-arts.
    \noindent{\textbf{Social GAN}}~\cite{gupta2018social} proposes a GAN-based method to predict diverse and socially acceptable trajectories.
       \noindent{\textbf{Desire}}~\cite{lee2017desire} uses CVAE to generate trajectory samples, which are then recursively ranked and refined.
    \noindent{\textbf{Multiverse}}~\cite{liang2020garden} selects multiple coarse trajectories from predicted OGMs by beam search and then refines them with continuous displacement vectors.
        \noindent{\textbf{SimAug}}~\cite{liang2020simaug} improves Multiverse~\cite{liang2020garden}'s robustness by utilizing simulated multi-view data. 
        \noindent{\textbf{P2T}}~\cite{deo2020trajectory} predicts the future trajectory conditioned on a plan generated by the deep MaxEnt IRL. 
        \noindent{\textbf{PECNet}}~\cite{mangalam2020not} is a goal-conditioned model dividing the task into goal estimation and trajectory prediction. 
        \noindent{\textbf{LB-EBM}}~\cite{pang2021trajectory} infers intermediate waypoints using latent vector sampled from a cost-based history.
        \noindent{\textbf{Y-Net}}~\cite{mangalam2021goals} models the future position's multimodality with heatmaps and samples a trajectory from the heatmap conditioned on sampled goal and waypoints. 
        \noindent{\textbf{V}}~\cite{wong2021view} is a concurrent method proposing a two-stage Transformer network to model the trajectory and its Fourier spectrum in the keypoints and interactions levels, respectively.

The performance of our model in the short-term trajectory prediction compared with state-of-the-art methods on the SDD is reported in \cref{tab:result}. $\operatorname{minADE}_{20}$ and $\operatorname{minFDE}_{20}$ values follow the original papers, while the offroad rates are computed using the released codes and models of different approaches.
On both data splits, our model achieves the best performance according to the metrics of $\operatorname{minADE}_{20}$ and $\operatorname{minADE}_{20}$ and offroad rate. Notably, our results are achieved without manually labeled semantic maps in Y-Net~\cite{mangalam2021goals} or simulation data in SimAug~\cite{liang2020simaug}. 

We also report our long-term prediction results on the inD in \cref{tab:inD}. Our results are again achieved without the manually annotated semantic maps in Y-net~\cite{mangalam2021goals}.

A set of qualitative examples are presented in the supplementary material, which demonstrates that our models are able to learn a diverse and feasible distribution and predict diverse representative trajectories.

\begin{table}
    \centering
    % \footnotesize
    \resizebox{0.7\linewidth}{!}{
    % \footnotesize
	\begin{tabular}{l|cc}
    \toprule
	 Model &  $\operatorname{minADE}_{20}$ & $\operatorname{minFDE}_{20}$  \\
	\midrule
	S-GAN~\cite{gupta2018social} &  38.57 & 84.61  \\
    PECNet~\cite{mangalam2020not} & 20.25 & 32.95  \\
    Y-net~\cite{mangalam2021goals} & 14.99 & 21.13 \\
    \midrule
    Ours & \textbf{13.09} & \textbf{19.39} \\
	\bottomrule
	\end{tabular}}
	\vspace{-0.2cm}
	\caption{Results on the inD in the long-term 30-second prediction.}
	\label{tab:inD}
\end{table}

\subsection{Ablation Study}
Ablation experiments on \textit{TrajNet} split are used to expose the significance of different components of our model:

\noindent{\textbf{OGMs decoder:}} First, we consider the ground-truth distribution approximation using one OGM obtained by a CNN acting on scene and motion features as R2P2~\cite{rhinehart2018r2p2}. Then, we study how effective our deconvolution parameterization in the ConvLSTM is for OGMs prediction. We implement three baseline OGMs prediction networks inspired by~\cite{ridel2020scene,jain2020discrete,park2020diverse}: $\operatorname{ConvLSTM}$ directly outputs OGMs from hidden maps; $\operatorname{ConvLSTM}$+DiscreteResidualFlow outputs the log-probability residual between OGMs from hidden maps; $\operatorname{ConvLSTM}$+GraphAttentionNetwork processes hidden maps with a graph attention network at each step. We train these models as our first training step. Results in~\cref{tab:OGM} about the OGM decoding loss measured by the NLL loss show that our approximation with different OGMs using deconvolution parameterization is the most effective. 

\begin{table}
    \centering
    \resizebox{0.9\linewidth}{!}{
    % \footnotesize
	\begin{tabular}{l|c}
    \toprule
	OGMs Prediction Model & $\mathcal{H}(p,\mathbf{O})$\\
	\midrule
	$\operatorname{CNN}$~\cite{rhinehart2018r2p2} & 17.52 \\
    $\operatorname{ConvLSTM}$~\cite{ridel2020scene} & 10.52	 \\
    $\operatorname{ConvLSTM}$+DiscreteResidualFlow~\cite{jain2020discrete} & 10.64 \\
    $\operatorname{ConvLSTM}$+GraphAttentionNetwork~\cite{park2020diverse} & 10.40 \\ 
    $\operatorname{ConvLSTM}$+Deconvolution (Ours) & \textbf{10.31}\\
	\bottomrule
	\end{tabular}}
	\vspace{-0.2cm}
	\caption{Comparison of four baselines and our method with close parameter numbers in predicting OGMs.}
	\label{tab:OGM}
\end{table}

\noindent{\textbf{Hyperparameter $\beta$:}} To investigate how the hyperparameter $\beta$ in the symmetric cross-entropy loss affects learning trajectory distribution, we train the policy network and the trajectory decoder with various $\beta$ values. To measure the learned distribution diversity, we leverage the $\operatorname{RF}_K$ metric from~\cite{park2020diverse}, \ie the ratio of the average FDE to the minimum FDE among $K$ predictions (${\operatorname{avgFDE}_K}/{\operatorname{minFDE}_K}$). A large ${\operatorname{avgFDE}_K}$ implies that predictions spread out while a small ${\operatorname{minFDE}_K}$ ensures predictions not being arbitrarily stochastic. The off-road rate metric is also applied to evaluate the distribution's precision. As shown in \cref{tab:beta}, with increasing $\beta$ values, the off-road rate and reverse cross-entropy decrease, implying a more precise distribution model, while the forward cross-entropy increases and $\operatorname{RF}_{20}$ decreases, meaning the distribution is getting less diverse. It shows that the hyperparameter $\beta$ can balance the predicted distribution's diversity and accuracy while a distribution only minimizing the forward cross-entropy can cover the data well but will produce implausible samples.

\begin{table}
    \centering
    % \footnotesize
    \resizebox{1\linewidth}{!}{
	\begin{tabular}{l|c|cccc}
    \toprule
	 Model & $\beta$ & $\mathcal{H}(p,q_\theta)$ & $\mathcal{H}(q_\theta,\mathbf{O})$  & $\operatorname{RF}_{20}$ & Offroad Rate \\
	\midrule
	Ours & 0 & \textbf{-32.19} & 24.13 & \textbf{7.52} & 0.035 \\
	Ours & 0.1 & -31.95 & 8.56  & 4.99 & 0.034 \\
	Ours & 0.2 & -31.75 & 7.536 & 4.23 & 0.030 \\
    Ours & 1 &  -31.52 & 4.21 & 2.70  & 0.022\\
    Ours & 10 &  -29.76 & \textbf{3.52} & 2.13  & \textbf{0.020}\\
	\midrule
	BC & 0.2 & -31.62 & 8.10 & 4.80 & 0.034 \\
	SR & 0.2 & -31.72 & 9.52 & 6.95 & 0.033 \\
	\bottomrule
	\end{tabular}}
	\vspace{-0.2cm}
	\caption{Effect of $\beta$ and reward in trajectory distribution forecasting based on the predicted OGMs.}
	\label{tab:beta}
\end{table}
% Since $\mathcal{H}(p,q_\theta)$ can be unbounded below, we resolve this by computing $-\mathbb{E}_{\eta \sim \mathcal{N}(0, \epsilon I)} \mathbb{E}_{x \sim p} \log q(x+\eta)$ for $\eta=0.001$ as~\cite{rhinehart2018r2p2}.
% 		\midrule
% 	No-GS & 1 & -31.19 & 6.26 & 3.97 & 0.027\\

\begin{table}
    \centering
    % \footnotesize
    \resizebox{1\linewidth}{!}{
	\begin{tabular}{l|ccc}
    \toprule
	 Method &  $\operatorname{minADE}_{20}$ & $\operatorname{minFDE}_{20}$ & Offroad Rate \\
	\midrule
	w/o refinement &  8.78 & 14.34 & \textbf{0.045}\\
    K-means & 7.64	 &  12.12 & 0.058 \\
    \midrule
    w/o end-to-end &  7.36 & 11.51 & 0.077 \\
    multi-task  & 6.94   & 10.68 & 0.066 \\
    variety loss & 8.16 & 13.04 & 0.084 \\
    Ours & \textbf{6.77} & \textbf{10.46} & 0.066 \\
	\bottomrule
	\end{tabular}}
	\vspace{-0.2cm}
	\caption{Effect of the refinement and training methods in predicting representative trajectories based on the learned trajectory distribution with $\beta=0.2$.}
	\label{tab:refine}
\end{table}

\noindent{\textbf{Reward layers:}} First, we study how the IRL learning method is beneficial compared to behavior cloning (BC). In the BC method, we ablate the value iteration network and directly output the non-stationary policy in replace of the non-stationary reward. Then, our non-stationary reward is compared with the stationary reward (SR) used in previous works~\cite{pflueger2019rover,deo2020trajectory}. The SR method is implemented by mapping the concatenation of the motion and scene feature maps into one reward map through two fully connected convolutional layers. Results in \cref{tab:beta} show that a non-stationary reward outperforms no reward or a stationary one in terms of forward and reverse cross-entropy and off-road rate.

\noindent{\textbf{Refinement network:}}
We consider two models without the refinement network. One removes the refinement network from our method. The other replaces our refinement network with K-means in~\cite{deo2020trajectory} and outputs $K$ cluster centers of trajectory samples as representatives. Both models are trained end-to-end using the variety loss based on the pre-trained distribution model. The comparison between \cref{tab:refine} and \cref{tab:result} bottom part shows that the refinement network is indispensable and more effective than K-means while increasing the offroad rate due to the diversity loss.

\noindent{\textbf{Training process:}} Firstly, we show the result of only completing the first three training processes without the end-to-end fine-tuning process. Besides, we also consider two other training processes. One is to train the network with the sum of all losses like multi-task. The other one is with the variety loss only. \cref{tab:refine} demonstrates that our training process with end-to-end fine-tuning can improve performance in predicting representative trajectories. We find that training only with variety loss is unstable and may not converge.

\section{Conclusion}
\label{sec:conclusion}

We have proposed an end-to-end interpretable trajectory distribution prediction model based on a grid plan. Our model can learn to produce a diverse and admissible trajectory distribution by minimizing the symmetric cross-entropy loss. We also design a flexible refinement network to generate a small set of representative trajectories. Finally, we demonstrate the effectiveness of our approach in two real-world datasets with state-of-the-art performance.

\clearpage
{\small
\bibliographystyle{ieee_fullname}
\bibliography{mybib}

\begin{thebibliography}{10}\itemsep=-1pt

\bibitem{alahi2016social}
Alexandre Alahi, Kratarth Goel, Vignesh Ramanathan, Alexandre Robicquet, Li
  Fei-Fei, and Silvio Savarese.
\newblock {Social LSTM}: Human trajectory prediction in crowded spaces.
\newblock In {\em CVPR}, pages 961--971, 2016.

\bibitem{amirian2019social}
Javad Amirian, Jean-Bernard Hayet, and Julien Pettr{\'e}.
\newblock {Social Ways}: Learning multi-modal distributions of pedestrian
  trajectories with gans.
\newblock In {\em CVPRW}, pages 0--0, 2019.

\bibitem{inDdataset}
Julian Bock, Robert Krajewski, Tobias Moers, Steffen Runde, Lennart Vater, and
  Lutz Eckstein.
\newblock {The inD Dataset}: A drone dataset of naturalistic road user
  trajectories at german intersections.
\newblock 2019.

\bibitem{casas2021mp3}
Sergio Casas, Abbas Sadat, and Raquel Urtasun.
\newblock {MP3}: A unified model to map, perceive, predict and plan.
\newblock In {\em CVPR}, pages 14403--14412, 2021.

\bibitem{dendorfer2020goal}
Patrick Dendorfer, Aljosa Osep, and Laura Leal-Taix{\'e}.
\newblock {Goal-GAN}: Multimodal trajectory prediction based on goal position
  estimation.
\newblock In {\em ACCV}, 2020.

\bibitem{deo2020trajectory}
Nachiket Deo and Mohan~M Trivedi.
\newblock Trajectory forecasts in unknown environments conditioned on
  grid-based plans.
\newblock {\em arXiv preprint arXiv:2001.00735}, 2020.

\bibitem{deo2021multimodal}
Nachiket Deo, Eric~M Wolff, and Oscar Beijbom.
\newblock Multimodal trajectory prediction conditioned on lane-graph
  traversals.
\newblock {\em arXiv preprint arXiv:2106.15004}, 2021.

\bibitem{eiffert2020probabilistic}
Stuart Eiffert, Kunming Li, Mao Shan, Stewart Worrall, Salah Sukkarieh, and
  Eduardo Nebot.
\newblock Probabilistic crowd gan: Multimodal pedestrian trajectory prediction
  using a graph vehicle-pedestrian attention network.
\newblock {\em IEEE Robotics and Automation Letters}, 5(4):5026--5033, 2020.

\bibitem{gupta2018social}
Agrim Gupta, Justin Johnson, Li Fei-Fei, Silvio Savarese, and Alexandre Alahi.
\newblock {Social GAN}: Socially acceptable trajectories with generative
  adversarial networks.
\newblock In {\em CVPR}, pages 2255--2264, 2018.

\bibitem{he2016deep}
Kaiming He, Xiangyu Zhang, Shaoqing Ren, and Jian Sun.
\newblock Deep residual learning for image recognition.
\newblock In {\em CVPR}, pages 770--778, 2016.

\bibitem{hu2020collaborative}
Yue Hu, Siheng Chen, Ya Zhang, and Xiao Gu.
\newblock Collaborative motion prediction via neural motion message passing.
\newblock In {\em CVPR}, pages 6319--6328, 2020.

\bibitem{huang2020diversitygan}
Xin Huang, Stephen~G McGill, Jonathan~A DeCastro, Luke Fletcher, John~J
  Leonard, Brian~C Williams, and Guy Rosman.
\newblock {DiversityGAN}: Diversity-aware vehicle motion prediction via latent
  semantic sampling.
\newblock {\em IEEE Robotics and Automation Letters}, 5(4):5089--5096, 2020.

\bibitem{huang2019stgat}
Yingfan Huang, HuiKun Bi, Zhaoxin Li, Tianlu Mao, and Zhaoqi Wang.
\newblock {STGAT}: Modeling spatial-temporal interactions for human trajectory
  prediction.
\newblock In {\em ICCV}, pages 6272--6281, 2019.

\bibitem{jain2020discrete}
Ajay Jain, Sergio Casas, Renjie Liao, Yuwen Xiong, Song Feng, Sean Segal, and
  Raquel Urtasun.
\newblock Discrete residual flow for probabilistic pedestrian behavior
  prediction.
\newblock In {\em Conference on Robot Learning}, pages 407--419, 2020.

\bibitem{jang2017categorical}
Eric Jang, Shixiang Gu, and Ben Poole.
\newblock Categorical reparameterization with gumbel-softmax.
\newblock In {\em ICLR}, 2017.

\bibitem{kim2017probabilistic}
ByeoungDo Kim, Chang~Mook Kang, Jaekyum Kim, Seung~Hi Lee, Chung~Choo Chung,
  and Jun~Won Choi.
\newblock Probabilistic vehicle trajectory prediction over occupancy grid map
  via recurrent neural network.
\newblock In {\em International Conference on Intelligent Transportation
  Systems}, pages 399--404, 2017.

\bibitem{Kingma2014Auto}
Diederik~P Kingma and Max Welling.
\newblock Auto-encoding variational bayes.
\newblock In {\em ICLR}, 2014.

\bibitem{kothari2021human}
Parth Kothari, Sven Kreiss, and Alexandre Alahi.
\newblock Human trajectory forecasting in crowds: A deep learning perspective.
\newblock {\em IEEE Transactions on Intelligent Transportation Systems}, 2021.

\bibitem{lee2017desire}
Namhoon Lee, Wongun Choi, Paul Vernaza, Christopher~B Choy, Philip~HS Torr, and
  Manmohan Chandraker.
\newblock Desire: Distant future prediction in dynamic scenes with interacting
  agents.
\newblock In {\em CVPR}, pages 336--345, 2017.

\bibitem{leung2016distributional}
Karen Leung, Edward Schmerling, and Marco Pavone.
\newblock Distributional prediction of human driving behaviours using mixture
  density networks.
\newblock {\em Technical report, Stanford University}, 2016.

\bibitem{li2017pedestrian}
Yuke Li.
\newblock Pedestrian path forecasting in crowd: A deep spatio-temporal
  perspective.
\newblock In {\em ACMMM}, pages 235--243, 2017.

\bibitem{liang2020simaug}
Junwei Liang, Lu Jiang, and Alexander Hauptmann.
\newblock {SimAug}: Learning robust representations from simulation for
  trajectory prediction.
\newblock In {\em ECCV}, pages 275--292, 2020.

\bibitem{liang2020garden}
Junwei Liang, Lu Jiang, Kevin Murphy, Ting Yu, and Alexander Hauptmann.
\newblock The garden of forking paths: Towards multi-future trajectory
  prediction.
\newblock In {\em CVPR}, pages 10508--10518, 2020.

\bibitem{luo2021safety}
Katie Luo, Sergio Casas, Renjie Liao, Xinchen Yan, Yuwen Xiong, Wenyuan Zeng,
  and Raquel Urtasun.
\newblock Safety-oriented pedestrian occupancy forecasting.
\newblock In {\em International Conference on Intelligent Robots and Systems},
  pages 1015--1022, 2021.

\bibitem{luo2018fast}
Wenjie Luo, Bin Yang, and Raquel Urtasun.
\newblock Fast and furious: Real time end-to-end 3d detection, tracking and
  motion forecasting with a single convolutional net.
\newblock In {\em CVPR}, pages 3569--3577, 2018.

\bibitem{makansi2019overcoming}
Osama Makansi, Eddy Ilg, Ozgun Cicek, and Thomas Brox.
\newblock Overcoming limitations of mixture density networks: A sampling and
  fitting framework for multimodal future prediction.
\newblock In {\em CVPR}, pages 7144--7153, 2019.

\bibitem{mangalam2021goals}
Karttikeya Mangalam, Yang An, Harshayu Girase, and Jitendra Malik.
\newblock From goals, waypoints \& paths to long term human trajectory
  forecasting.
\newblock In {\em ICCV}, 2021.

\bibitem{mangalam2020not}
Karttikeya Mangalam, Harshayu Girase, Shreyas Agarwal, Kuan-Hui Lee, Ehsan
  Adeli, Jitendra Malik, and Adrien Gaidon.
\newblock It is not the journey but the destination: Endpoint conditioned
  trajectory prediction.
\newblock In {\em ECCV}, pages 759--776, 2020.

\bibitem{marchetti2020mantra}
Francesco Marchetti, Federico Becattini, Lorenzo Seidenari, and Alberto~Del
  Bimbo.
\newblock Mantra: Memory augmented networks for multiple trajectory prediction.
\newblock In {\em CVPR}, pages 7143--7152, 2020.

\bibitem{mercat2020multi}
Jean Mercat, Thomas Gilles, Nicole El~Zoghby, Guillaume Sandou, Dominique
  Beauvois, and Guillermo~Pita Gil.
\newblock Multi-head attention for multi-modal joint vehicle motion
  forecasting.
\newblock In {\em International Conference on Robotics and Automation}, pages
  9638--9644, 2020.

\bibitem{minoura2020crowd}
Hiroaki Minoura, Ryo Yonetani, Mai Nishimura, and Yoshitaka Ushiku.
\newblock Crowd density forecasting by modeling patch-based dynamics.
\newblock {\em IEEE Robotics and Automation Letters}, 6(2):287--294, 2020.

\bibitem{Mohajerin2019MultiStep}
Nima Mohajerin and Mohsen Rohani.
\newblock Multi-step prediction of occupancy grid maps with recurrent neural
  networks.
\newblock {\em CVPR}, pages 10592--10600, 2019.

\bibitem{pang2021trajectory}
Bo Pang, Tianyang Zhao, Xu Xie, and Ying~Nian Wu.
\newblock Trajectory prediction with latent belief energy-based model.
\newblock In {\em CVPR}, pages 11814--11824, 2021.

\bibitem{park2020diverse}
Seong~Hyeon Park, Gyubok Lee, Jimin Seo, Manoj Bhat, Minseok Kang, Jonathan
  Francis, Ashwin Jadhav, Paul~Pu Liang, and Louis-Philippe Morency.
\newblock Diverse and admissible trajectory forecasting through multimodal
  context understanding.
\newblock In {\em ECCV}, pages 282--298, 2020.

\bibitem{pflueger2019rover}
Max Pflueger, Ali Agha, and Gaurav~S Sukhatme.
\newblock {Rover-IRL}: Inverse reinforcement learning with soft value iteration
  networks for planetary rover path planning.
\newblock {\em IEEE Robotics and Automation Letters}, 4(2):1387--1394, 2019.

\bibitem{phan2020covernet}
Tung Phan-Minh, Elena~Corina Grigore, Freddy~A Boulton, Oscar Beijbom, and
  Eric~M Wolff.
\newblock {CoverNet}: Multimodal behavior prediction using trajectory sets.
\newblock In {\em CVPR}, pages 14074--14083, 2020.

\bibitem{rehder2018pedestrian}
Eike Rehder, Florian Wirth, Martin Lauer, and Christoph Stiller.
\newblock Pedestrian prediction by planning using deep neural networks.
\newblock In {\em International Conference on Robotics and Automation}, pages
  5903--5908, 2018.

\bibitem{rhinehart2018r2p2}
Nicholas Rhinehart, Kris~M Kitani, and Paul Vernaza.
\newblock R2p2: A reparameterized pushforward policy for diverse, precise
  generative path forecasting.
\newblock In {\em ECCV}, pages 772--788, 2018.

\bibitem{ridel2020scene}
Daniela Ridel, Nachiket Deo, Denis Wolf, and Mohan Trivedi.
\newblock Scene compliant trajectory forecast with agent-centric
  spatio-temporal grids.
\newblock {\em IEEE Robotics and Automation Letters}, 5(2):2816--2823, 2020.

\bibitem{robicquet2016learning}
Alexandre Robicquet, Amir Sadeghian, Alexandre Alahi, and Silvio Savarese.
\newblock Learning social etiquette: Human trajectory understanding in crowded
  scenes.
\newblock In {\em ECCV}, pages 549--565, 2016.

\bibitem{rudenko2020human}
Andrey Rudenko, Luigi Palmieri, Michael Herman, Kris~M Kitani, Dariu~M Gavrila,
  and Kai~O Arras.
\newblock Human motion trajectory prediction: A survey.
\newblock {\em The International Journal of Robotics Research}, 39(8):895--935,
  2020.

\bibitem{Sadeghian2018Trajnet}
Amir Sadeghian, Vineet Kosaraju, Agrim Gupta, Silvio Savarese, and A Alahi.
\newblock Trajnet: Towards a benchmark for human trajectory prediction.
\newblock {\em arXiv preprint}, 2018.

\bibitem{sadeghian2019sophie}
Amir Sadeghian, Vineet Kosaraju, Ali Sadeghian, Noriaki Hirose, Hamid
  Rezatofighi, and Silvio Savarese.
\newblock Sophie: An attentive gan for predicting paths compliant to social and
  physical constraints.
\newblock In {\em CVPR}, pages 1349--1358, 2019.

\bibitem{salzmann2020trajectron++}
Tim Salzmann, Boris Ivanovic, Punarjay Chakravarty, and Marco Pavone.
\newblock Trajectron++: Dynamically-feasible trajectory forecasting with
  heterogeneous data.
\newblock In {\em ECCV}, pages 683--700, 2020.

\bibitem{2016Value}
Aviv Tamar, Yi Wu, Garrett Thomas, Sergey Levine, and Pieter Abbeel.
\newblock Value iteration networks.
\newblock In {\em NeurIPS}, pages 2154--2162, 2016.

\bibitem{thiede2019analyzing}
Luca~Anthony Thiede and Pratik~Prabhanjan Brahma.
\newblock Analyzing the variety loss in the context of probabilistic trajectory
  prediction.
\newblock In {\em ICCV}, pages 9954--9963, 2019.

\bibitem{vaswani2017attention}
Ashish Vaswani, Noam Shazeer, Niki Parmar, Jakob Uszkoreit, Llion Jones,
  Aidan~N Gomez, {\L}ukasz Kaiser, and Illia Polosukhin.
\newblock Attention is all you need.
\newblock In {\em NeurIPS}, pages 5998--6008, 2017.

\bibitem{wang2019symmetric}
Yisen Wang, Xingjun Ma, Zaiyi Chen, Yuan Luo, Jinfeng Yi, and James Bailey.
\newblock Symmetric cross entropy for robust learning with noisy labels.
\newblock In {\em ICCV}, pages 322--330, 2019.

\bibitem{wong2021view}
Conghao Wong, Beihao Xia, Ziming Hong, Qinmu Peng, and Xinge You.
\newblock {View Vertically}: A hierarchical network for trajectory prediction
  via fourier spectrums.
\newblock {\em arXiv preprint arXiv:2110.07288}, 2021.

\bibitem{wulfmeier2017large}
Markus Wulfmeier, Dushyant Rao, Dominic~Zeng Wang, Peter Ondruska, and Ingmar
  Posner.
\newblock Large-scale cost function learning for path planning using deep
  inverse reinforcement learning.
\newblock {\em The International Journal of Robotics Research},
  36(10):1073--1087, 2017.

\bibitem{xingjian2015convolutional}
Shi Xingjian, Zhourong Chen, Hao Wang, Dit-Yan Yeung, Wai-Kin Wong, and
  Wang-chun Woo.
\newblock {Convolutional LSTM network}: A machine learning approach for
  precipitation nowcasting.
\newblock In {\em NeurIPS}, pages 802--810, 2015.

\bibitem{yuan2020diverse}
Ye Yuan and Kris Kitani.
\newblock Diverse trajectory forecasting with determinantal point processes.
\newblock In {\em ICLR}, 2020.

\bibitem{zeng2019end}
Wenyuan Zeng, Wenjie Luo, Simon Suo, Abbas Sadat, Bin Yang, Sergio Casas, and
  Raquel Urtasun.
\newblock End-to-end interpretable neural motion planner.
\newblock In {\em CVPR}, pages 8660--8669, 2019.

\bibitem{zhang2019sr}
Pu Zhang, Wanli Ouyang, Pengfei Zhang, Jianru Xue, and Nanning Zheng.
\newblock {SR-LSTM}: State refinement for lstm towards pedestrian trajectory
  prediction.
\newblock In {\em CVPR}, pages 12085--12094, 2019.

\bibitem{zhao2020tnt}
Hang Zhao, Jiyang Gao, Tian Lan, Chen Sun, Benjamin Sapp, Balakrishnan
  Varadarajan, Yue Shen, Yi Shen, Yuning Chai, Cordelia Schmid, Congcong Li,
  and Dragomir Anguelov.
\newblock {TNT}: Target-driven trajectory prediction.
\newblock In {\em Conference on Robot Learning}, 2020.

\bibitem{zhao2021you}
He Zhao and Richard~P Wildes.
\newblock Where are you heading? dynamic trajectory prediction with expert goal
  examples.
\newblock In {\em ICCV}, pages 7629--7638, 2021.

\bibitem{zhou2021informer}
Haoyi Zhou, Shanghang Zhang, Jieqi Peng, Shuai Zhang, Jianxin Li, Hui Xiong,
  and Wancai Zhang.
\newblock Informer: Beyond efficient transformer for long sequence time-series
  forecasting.
\newblock In {\em AAAI}, pages 11106--11115, 2021.

\bibitem{ziebart2008maximum}
Brian~D Ziebart, Andrew~L Maas, J~Andrew Bagnell, and Anind~K Dey.
\newblock Maximum entropy inverse reinforcement learning.
\newblock In {\em AAAI}, volume~8, pages 1433--1438, 2008.

\end{thebibliography}
}

\end{document}

\title{End-to-End Trajectory Distribution Prediction Based on Occupancy Grid Maps (Supplementary Material)} 
\maketitle

\section{Occupancy Grid Maps}
\begin{figure}[htbp]
	\centering
	\includegraphics[width=\linewidth]{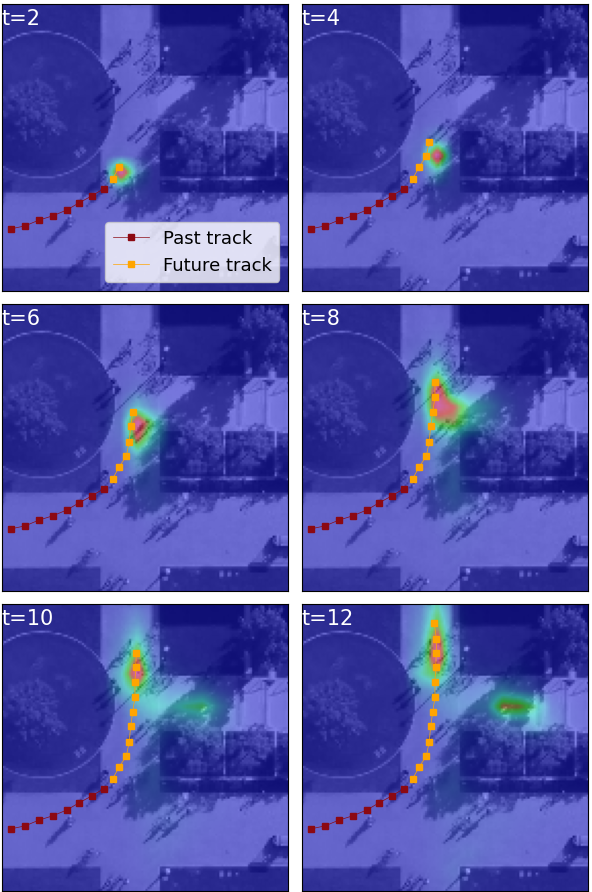}
	\vspace{-0.6cm}
	\caption{OGMs prediction results on the SDD. Warmer color indicates higher occupancy probability, while colder color represents lower occupancy probability.}
	\vspace{-0.65cm}
	\label{fig:sdd_ogm}
\end{figure}

\begin{figure}[htbp]
	\centering
	\includegraphics[width=\linewidth]{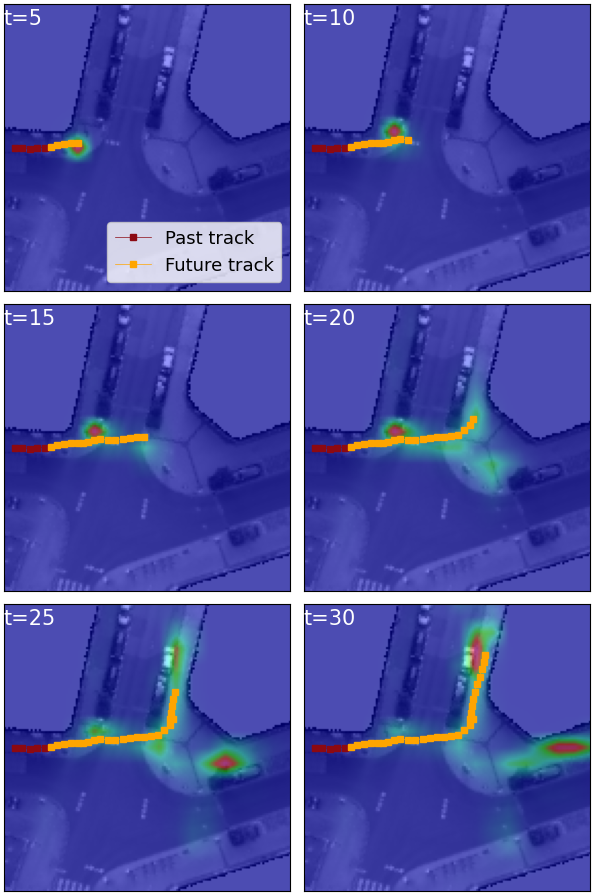}
	\vspace{-0.5cm}
	\caption{OGMs prediction results on the inD. }
	\vspace{-0.4cm}
	\label{fig:ind_ogm}
\end{figure}

In this section, we show examples of occupancy grid maps (OGMs) prediction results on the two datasets: Stanford Drone Dataset (SDD) and Intersection Drone Dataset (inD). Our temporal OGMs represent each future position distribution with an OGM, so we show the predicted OGMs from our model at 6 time steps. \cref{fig:sdd_ogm} and \cref{fig:ind_ogm} show that our temporal OGMs can capture the scene compliant future position distribution while reflecting the multimodality.

\section{Reward Maps}

\begin{figure}[htbp]
	\centering
	\includegraphics[width=\linewidth]{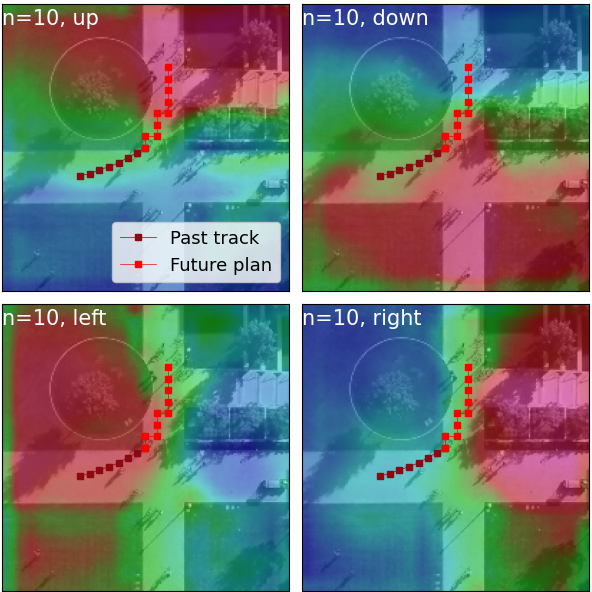}
	\vspace{-0.5cm}
	\caption{Reward maps of taking the 4 adjacent movements ${up, down, left, right}$ at MDP step $n=10$ on the SDD. Warmer color represents higher reward value, while colder color implies lower reward value.}
	\vspace{-0.4cm}
	\label{fig:sdd_reward}
\end{figure}

\begin{figure}[htbp]
	\centering
	\includegraphics[width=\linewidth]{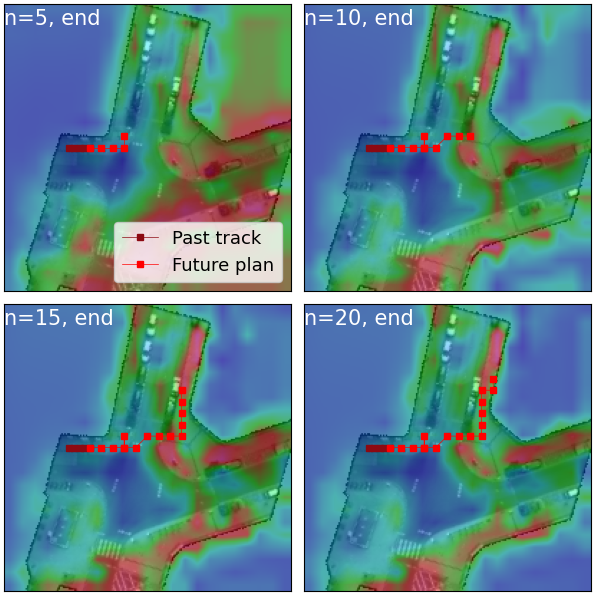}
	\vspace{-0.5cm}
	\caption{Reward maps of taking the \textit{end} action on the inD. }
	\vspace{-0.4cm}
	\label{fig:ind_reward}
\end{figure}

Our model learn non-stationary rewards dependent on the history trajectories and neighboring scene for producing non-stationary policy. Our non-stationary rewards  $\mathbf{r}^n:\mathcal{S} \times \mathcal{A} \rightarrow \mathbb{R}$ are dependent on the action and state, where the action set includes 4 adjacent movements ${up, down, left, right}$ and an $end$ action leading to the absorbing state. For simplicity, we show the reward maps of taking four moving actions at the fixed MDP steps $n=10$ on the SDD (see \cref{fig:sdd_reward}) while the $end$ action at four different MDP steps $n=5,10,15,20$ on the inD (see \cref{fig:ind_reward}). 
% an action has higher rewards on the roads of the corresponding direction than the grassland or the roads of inconsistent direction relative to the current position
As shown in \cref{fig:sdd_reward}, the regions with the highest rewards (\ie, in red color) are generally consistent with the action direction.
% for each action, there are higher rewards at locations in the action direction relative to the current position.
\cref{fig:ind_reward} illustrates that the roadside or crosswalk regions have higher $end$ rewards than the car lane and impassable regions,
% there are higher $end$ rewards on the roadside or crosswalk instead of the road or the regions out of the scene. 
which is reasonable since the used training data of inD are pedestrian trajectories instead of vehicle tracks.

\section{Policy Maps}
\begin{figure}[t]
	\centering
	\includegraphics[width=\linewidth]{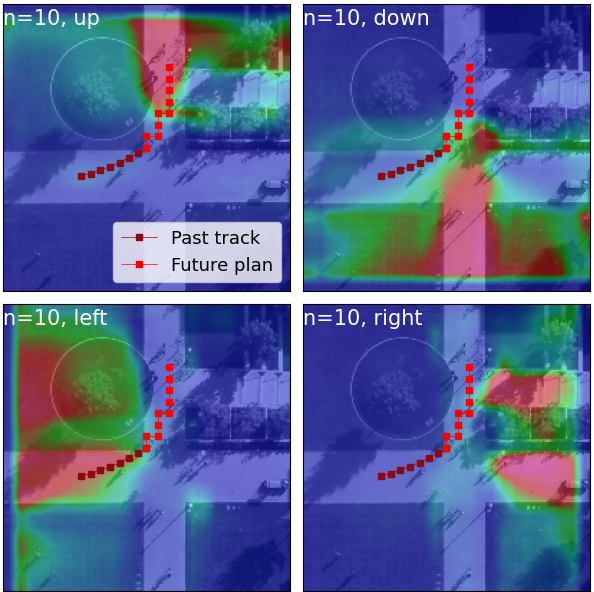}
	\vspace{-0.5cm}
	\caption{Policy maps of taking four movements ${up, down, left, right}$ at MDP step $n=10$ on the SDD. Warmer color means higher action probability, while colder color means lower action probability.}	
	\vspace{-0.4cm}
	\label{fig:sdd_policy}
\end{figure}

\begin{figure}[t]
	\centering
	\includegraphics[width=\linewidth]{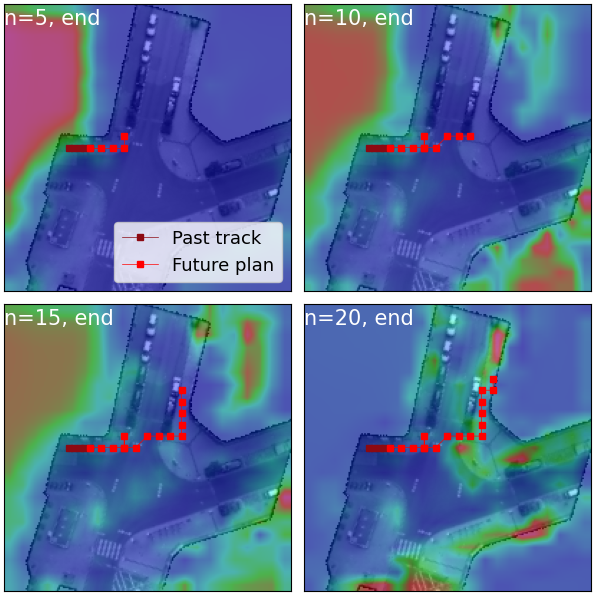}
	\vspace{-0.5cm}
	\caption{Policy maps of taking the \textit{end} action on the inD.}	
	\vspace{-0.4cm}
	\label{fig:ind_policy}
\end{figure}

We generate a non-stationary policy using an approximate value iteration network based on the non-stationary rewards. Similarly, we show the policy maps of taking four moving actions at the fixed MDP steps $n=10$ on the SDD (see \cref{fig:sdd_policy}) while the $end$ action at four different MDP steps $n=5,10,15,20$ on the inD (see \cref{fig:ind_policy}). 

% \cref{fig:sdd_policy} shows that it is more likely to take an action to move on the road in the action direction relative to the current position. \cref{fig:ind_policy} demonstrates that the $end$ action has a higher probability on the roadside and the regions out of the scene. Besides, there are higher probabilities to end on the roadside at the later steps to generate a longer plan. 
\cref{fig:sdd_policy} shows that our model learns to take an action to move on the road consistent with the action direction and avoid the terrains. \cref{fig:ind_policy} demonstrates that it learns to end at the roadsides in the last few MDP steps, or the impassable regions, where the pedestrian trajectories seen in the training data usually end in 30 seconds future.

\subsection{Plan-conditioned Trajectory}

\begin{figure}[htbp]
	\centering
	\includegraphics[width=\linewidth]{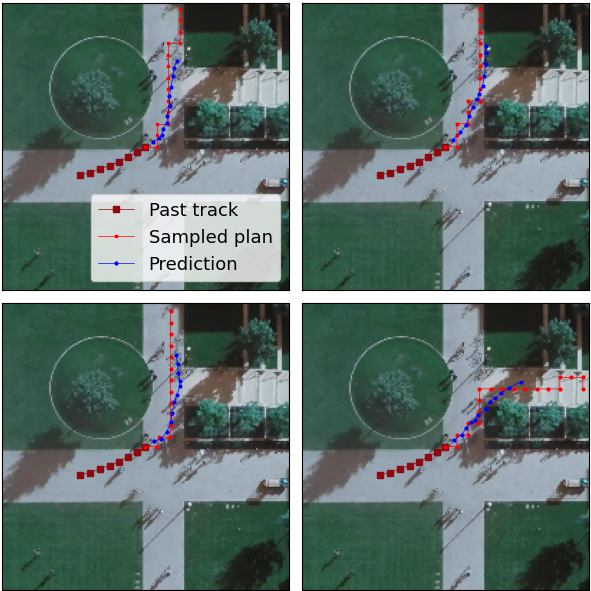}
	\vspace{-0.5cm}
	\caption{Four grid-based plans sampled from the non-stationary policy and the generated trajectories conditioned on the plans on the SDD.}	
	\vspace{-0.4cm}
	\label{fig:sdd_plan}
\end{figure}

\begin{figure}[htbp]
	\centering
	\includegraphics[width=\linewidth]{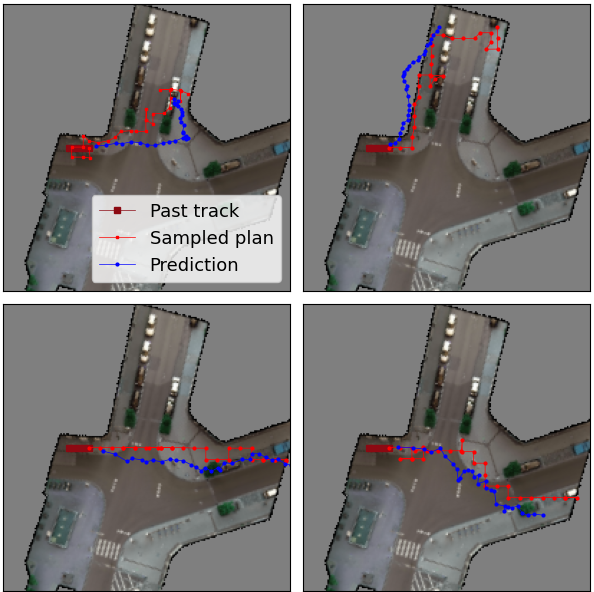}
	\vspace{-0.5cm}
	\caption{Four grid-based plans and corresponding trajectories on the inD.}	
	\vspace{-0.4cm}
	\label{fig:ind_plan}
\end{figure}

The grid-based plans are sampled from the non-stationary policy by Gumbel-Softmax trick and bilinearly interpolation. Then, we recursively generate a trajectory using an RNN with multi-head attention on the sampled plan. We show four examples of an sampled plan and its corresponding trajectory on SDD in \cref{fig:sdd_plan} and inD in \cref{fig:ind_plan}. We can observe that the sampled plans are scene compliant and the trajectory follows its corresponding plan while keeping smooth. 

\section{Trajectory Distribution}

\begin{figure}[htbp]
	\centering
	\includegraphics[width=\linewidth]{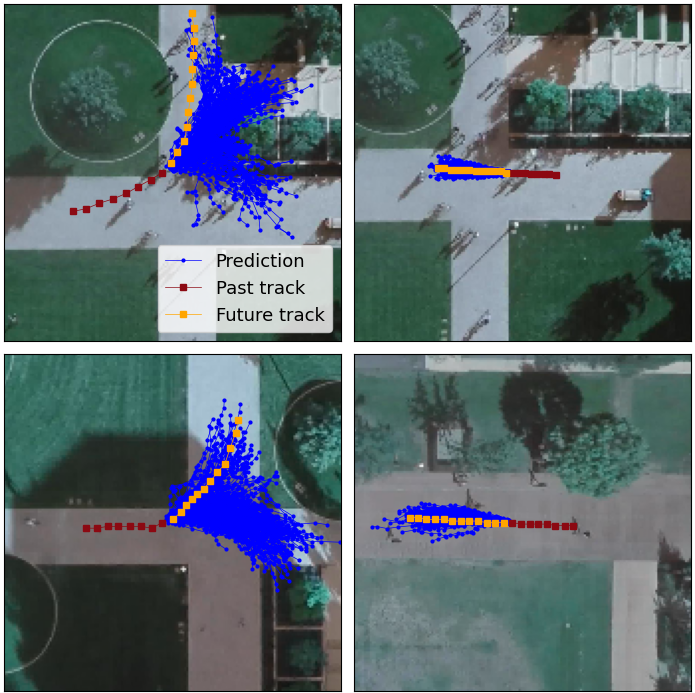}
	\vspace{-0.5cm}
	\caption{Predicted trajectory distributions on the SDD.}	
	\vspace{-0.4cm}
	\label{fig:sdd_dist}
\end{figure}

\begin{figure}[htbp]
	\centering
	\includegraphics[width=\linewidth]{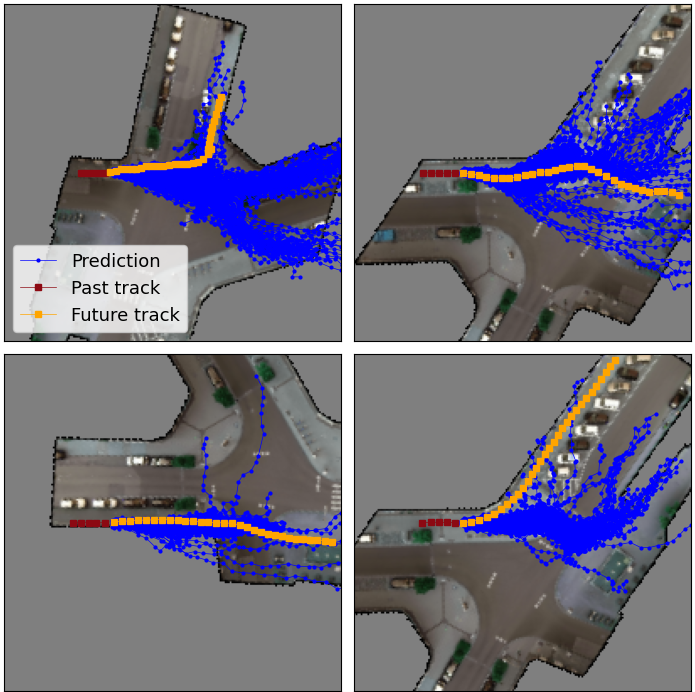}
	\vspace{-0.5cm}
	\caption{Predicted trajectory distributions on the inD.}	
	\vspace{-0.4cm}
	\label{fig:ind_dist}
\end{figure}

By minimizing the approximate symmetric cross-entropy loss, we learn a trajectory distribution close to the ground-truth. Examples of the predictive trajectory distribution from SDD and inD are shown in \cref{fig:sdd_dist} and \cref{fig:ind_dist}. We can see that our predicted trajectory distributions are diverse and feasible.
 
\section{Representative Trajectories}

\begin{figure}[htbp]
	\centering
	\includegraphics[width=\linewidth]{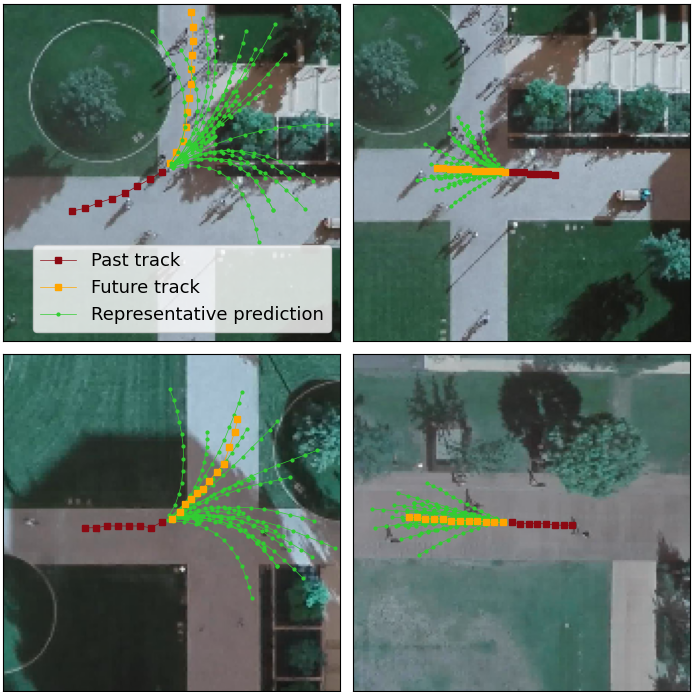}
	\vspace{-0.5cm}
	\caption{Representative trajectories predicted on the SDD.}	
	\vspace{-0.4cm}
	\label{fig:sdd_re}
\end{figure}

\begin{figure}[htbp]
	\centering
	\includegraphics[width=\linewidth]{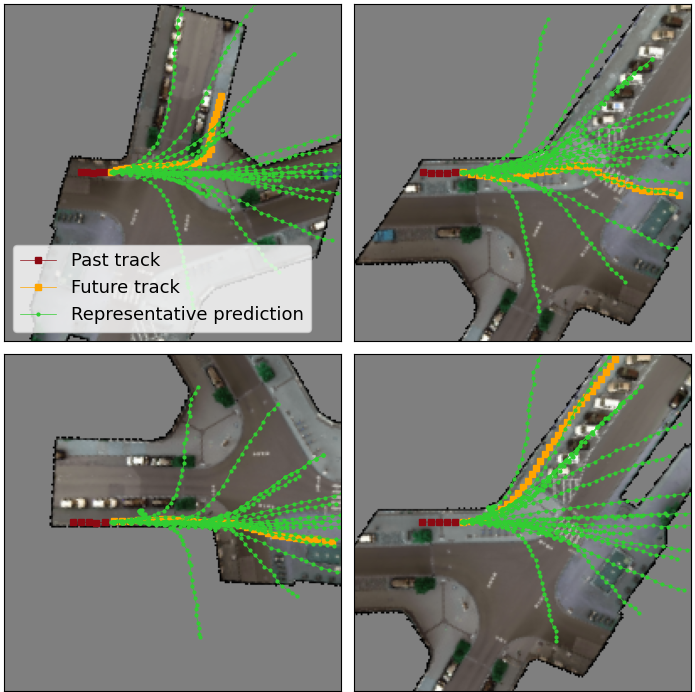}
	\vspace{-0.5cm}
	\caption{Representative trajectories predicted on the inD.}	
	\vspace{-0.4cm}
	\label{fig:ind_re}
\end{figure}

We design a Transformer-based refinement network to generate a small set of representative trajectories based on a large number of trajectories sampled from the trajectory distribution. \cref{fig:sdd_re} and \cref{fig:ind_re} show qualitative examples from SDD and inD. We note that the representative trajectories are diverse and capable to cover the ground-truth future trajectory due to the adopted variety loss.

